\documentclass[wcp]{jmlr}

\usepackage{amsmath}
\usepackage{amsfonts}
\usepackage{cancel}
\usepackage{caption}
\usepackage[section]{placeins}
\usepackage{float}
\usepackage{cleveref}

\usepackage{algorithm}
\usepackage{algorithmic}
\usepackage{graphicx}
\usepackage{amssymb}
\usepackage{url}
\usepackage{etoolbox}
\usepackage{cases}
\usepackage{bbm}
\usepackage{mathtools}
\usepackage{tikz}
\usetikzlibrary{matrix}

\usepackage{hyperref}
\hypersetup{
colorlinks=true,
 citecolor=blue,
 linkcolor=red,
 urlcolor=magenta}

\DeclareSymbolFont{extraup}{U}{zavm}{m}{n}
\DeclareMathSymbol{\varheart}{\mathalpha}{extraup}{86}
\def\fixme[#1]{\textcolor{red}{$\varheart$}\footnote{\textcolor{red}{#1}}}

\title[Particle Gibbs for iHMM]{A Linear-Time Particle Gibbs Sampler for Infinite Hidden Markov Models}

 \author{\Name{Nilesh Tripuraneni*} \Email{nt357@cam.ac.uk} \\
 \Name{Shane Gu*} \Email{sg717@cam.ac.uk} \\ 
 \Name{Hong Ge} \Email{hg344@cam.ac.uk} \\
 \Name{Zoubin Ghahramani} \Email{zoubin@eng.cam.ac.uk} \\
 \addr Department of Engineering, University of Cambridge, Trumpington Street,  Cambridge, CP2 1PZ}

\newenvironment{description2}
{ \begin{description}
    \setlength{\itemsep}{0pt}
    \setlength{\parskip}{0pt}
    \setlength{\parsep}{0pt}     }
{ \end{description}              }

\newcommand{\nwc}{\newcommand}
\nwc{\as}{\textrm{a.s.}}
\nwc{\defas}{:=}
\def\given{\,|\,}

\nwc{\dist}{\ \sim\ }
\nwc{\distiid}{\stackrel{\mathrm{iid}}{\sim}}

\newcommand{\rnwc}{\renewcommand}
\rnwc{\vec}[1]{\boldsymbol{\mathbf{#1}}}

\begin{document}

\maketitle

\begin{abstract}

Infinite Hidden Markov Models (iHMM's) are an attractive, nonparametric generalization of the classical Hidden Markov Model which can automatically infer the number of hidden states in the system.  However, due to the infinite-dimensional nature of transition dynamics performing inference in the iHMM is difficult. In this paper, we present an infinite-state Particle Gibbs (PG) algorithm to resample state trajectories for the iHMM. The proposed algorithm uses an efficient proposal optimized for iHMMs and leverages ancestor sampling to suppress degeneracy of the standard PG algorithm. Our algorithm demonstrates significant convergence improvements on synthetic and real world data sets. Additionally, the infinite-state PG algorithm has linear-time complexity in the number of states in the sampler, while competing methods scale quadratically. 

\end{abstract}
\begin{keywords}
Bayesian Nonparametrics, Hidden Markov Models, Model Selection, Particle MCMC
\end{keywords}

\section{Introduction}
\footnotetext{*equal contribution}
\label{sec:intro}

Hidden Markov Models (HMM's) are among the most widely adopted latent-variable models 
used to model time-series datasets in the statistics and machine learning communities. 
They have also been successfully applied in a variety of domains including genomics, language, and finance 
where sequential data naturally arises \citep{rabiner1989tutorial, bishop2006pattern}.

One possible disadvantage of the finite-state space HMM framework is that one must a-priori specify the number of latent states $K$. Standard model selection techniques can be applied to the finite state-space HMM but bear a high computational overhead since they require the repetitive training\big/exploration of many HMM's of different sizes. 

Bayesian nonparametric methods offer an attractive alternative to this problem by adapting their effective model complexity to fit the data. In particular, \citet{beal2001infinite} constructed an HMM over a countably infinite state-space using a Hierarchical Dirichlet Process (HDP) prior over the rows of the transition matrix. Various approaches have been taken to perform full posterior inference over the latent states, transition\big/emission distributions and hyperparameters since it is impossible to directly apply the forward-backwards algorithm due to the infinite-dimensional size of the state space. The original Gibbs sampling approach proposed in \citet{TehJorBea2006} suffered from slow mixing due to the strong correlations between nearby time steps often present in time-series data ~\citep{scott2002bayesian}. However, ~\citet{VanSaaTeh2008a} introduced a set of auxiliary slice variables to dynamically ``truncate" the state space to be finite (referred to as beam sampling), allowing them to use dynamic programming to jointly resample the latent states thus circumventing the problem. Despite the power of the beam-sampling scheme, ~\citet{fox2008hdp} found that application of the beam sampler to the (sticky) iHMM resulted in slow mixing relative to an inexact, blocked sampler due to the introduction of auxiliary slice variables in the sampler.

The main contributions of this paper are to derive an infinite-state PG algorithm for the iHMM using the stick-breaking construction for the HDP, and constructing an optimal importance proposal to efficiently resample its latent state trajectories. The proposed algorithm is compared to existing state-of-the-art inference algorithms for iHMMs, and empirical evidence suggests that the infinite-state PG algorithm consistently outperforms its alternatives. Furthermore, by construction the time complexity of the proposed algorithm is $\mathcal{O}(TNK)$.

Here $T$ denotes the length of the sequence, $N$ denotes the number of particles in the PG sampler, and $K$ denotes the number of ``active" states in the model.  Despite the simplicity of sampler, we find in a variety of synthetic and real-world experiments that these particle methods dramatically improve convergence of the sampler, while being more scalable.

We will first define the iHMM and sticky iHMM, reviewing the the Dirichlet Process (DP) and Hierarchical Dirichlet Process (HDP) in our appendix, in \Cref{sec:model-n-notation}. Then we move onto to the description of MCMC sampling scheme in \Cref{sec:inference}. In \Cref{sec:empirical} we present our results on a variety of synthetic and real-world datasets.

\section{Model and Notation}\label{sec:model-n-notation}
\begin{figure}[tb]
\begin{center}
\includegraphics[trim=0cm 6cm 0cm 6cm, clip, width=0.5\textwidth]{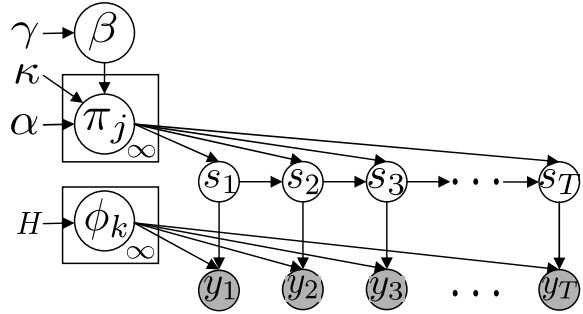}
\end{center}
\vspace{-0.6cm}
\caption{Graphical Model for the sticky HDP-HMM (setting $\kappa=0$ recovers the HDP-HMM)}
\end{figure}

\subsection{Infinite Hidden Markov Models}
We can formally define the iHMM (we review the theory of the HDP in our appendix) as follows:
\begin{align}\begin{split}
\vec{\beta} &\sim \text{GEM}(\gamma), \\
\vec{\pi}_{j} | \vec{\beta} &\distiid \text{DP}(\alpha, \vec{\beta}), \quad \phi_{j} \distiid H, \quad j=1,\ldots, \infty\\
 s_{t} | s_{t-1} &\sim \mathcal{C}at(\cdot | \vec{\pi}_{s_{t-1}}),  \quad y_{t} | s_{t} \sim f(\cdot | \phi_{s_{t}}), \quad t=1,\ldots, T.
\end{split}
\label{ihmm}
\end{align}
Here $\vec{\beta}$ is the shared DP measure defined on integers $\mathbb{Z}$. Here $s_{1:T} = (s_1, ..., s_{T})$ are the latent states of the iHMM, $y_{1:T} = (y_1, ..., y_T)$ are the observed data, and $\phi_{j}$ parametrizes the emission distribution $f$. Usually $H$ and $f$ are chosen to be conjugate to simplify the inference. $\beta_{k'}$ can be interpreted as the prior mean for transition probabilities into state $k'$, with $\alpha$ governing the variability of the prior mean across the rows of the transition matrix. The hyper-parameter $\gamma$ controls how concentrated or diffuse the probability mass of $\vec{\beta}$ will be over the states of the transition matrix. To connect the HDP with the iHMM, note that given a draw from the HDP $G_{k} = \sum_{k'=1}^{\infty} \vec{\pi}_{kk'} \delta_{\phi_{k'}}$ we identify $\vec{\pi}_{kk'}$ with the transition probability from state $k$ to state $k'$ where $\phi_{k'}$ parametrize the emission distributions.

Note that fixing $\vec{\beta}=(\frac{1}{K}, ...., \frac{1}{K},0,0...)$ implies only transitions between the first $K$ states of the transition matrix are ever possible, leaving us with the finite Bayesian HMM. If we define a finite, hierarchical Bayesian HMM by drawing
\begin{align}
\begin{split}
\vec{\beta} &\sim \mathcal{D}ir(\gamma/K, ..., \gamma/K)  \\
\vec{\pi}_k &\sim \mathcal{D}ir(\alpha \vec{\beta}) 
\label{eq:finite:hmm}
\end{split}
\end{align}
with joint density over the latent/hidden states as $$p_{\phi}(s_{1:T}, y_{1:T}) = \Pi_{t=1}^{T} \vec{\pi}(s_t | s_{t-1}) f_{\phi}(y_t | s_t)$$ then after taking $K \to \infty$, the hierarchical prior in \Cref{eq:finite:hmm} approaches the HDP. 

\subsection{Prior and Emission Distribution Specification}

The hyperparameter $\alpha$ governs the variability of the prior mean across the rows of the transition matrix and $\gamma$ controls how concentrated or diffuse the probability mass of $\vec{\beta}$ will be over the states of the transition matrix. However, in the HDP-HMM we have each row of the transition matrix is drawn as $\vec{\pi}_{j} \sim \text{DP}(\alpha, \vec{\beta})$. Thus the HDP prior doesn't differentiate self-transitions from jumps between different states. This can be especially problematic in the non-parametric setting, since non-Markovian state persistence in data can lead to the creation of unnecessary extra states and unrealistically, rapid switching dynamics in our model.  In \citet{fox2008hdp}, this problem is addressed by including a self-transition bias parameter into the distribution of transitioning probability vector $\vec{\pi}_{j}$:
\begin{equation}
\vec{\pi}_{j} \sim \text{DP}(\alpha+\kappa, \frac{\alpha \beta + \kappa \delta_{j}}{\alpha + \kappa} )
\end{equation}
to incorporate prior beliefs that smooth, state-persistent dynamics are more probable. Such a construction only involves the introduction of one further hyperparameter $\kappa$  which controls the ``stickiness" of the transition matrix (note a similar self-transition was explored in \citet{beal2001infinite}). 

For the standard iHMM, most approaches to inference have placed vague gamma hyper-priors on the hyper-parameters $\alpha$ and $\gamma$, which can be resampled efficiently as in \citet{TehJorBea2006}. Similarly in the sticky iHMM, in order to maintain tractable resampling of hyper-parameters \citet{fox2008hdp} chose to place vague gamma priors on $\gamma$, $\alpha +\kappa$, and a beta prior on $\kappa/(\alpha+\kappa)$. In this work we follow ~\citet{TehJorBea2006, fox2008hdp} and place priors $\gamma \sim \text{Gamma}(a_\gamma, b_\gamma)$, $\alpha + \kappa \sim \text{Gamma}(a_s, b_s)$, and $\kappa \sim \text{Beta}(a_\kappa, b_\kappa)$ on the hyper-parameters.

We consider two conjugate emission models for the output states of the iHMM -- a multinomial emission distribution for discrete data, and a normal emission distribution for continuous data. For discrete data we choose $\phi_k \sim \mathcal{D}ir(\alpha_\phi)$ with $f(\cdot\given \phi_{s_t}) = \mathcal{C}at(\cdot | \phi_k)$. For continuous data we choose $\phi_k = (\mu, \sigma^2) \sim \mathcal{N}IG(\mu, \lambda, \alpha_\phi, \beta_\phi)$ with $f(\cdot\given \phi_{s_t}) = \mathcal{N}(\cdot | \phi_k = (\mu, \sigma^2))$.

\section{Posterior Inference for the iHMM}\label{sec:inference}
Let us first recall the collection variables we need to sample: $\vec{\beta}$ is a shared DP base measure, $(\vec{\pi}_k)$ is the transition matrix acting on the latent states, while $\vec{\phi}_{k}$ parametrizes the emission distribution $f$, $k=1,\ldots,K$.   
We can then resample the variables of the iHMM in a series of Gibbs steps:

\vskip 0.1cm
\noindent\emph{Step 1: Sample $s_{1:T} \given y_{1:T}, \phi_{1:K}, \boldsymbol{\beta}, \vec{\pi}_{1:K}$.}\\
\noindent\emph{Step 2: Sample $\vec{\beta} \given s_{1:T}, \gamma$.}\\
\noindent\emph{Step 3: Sample $\vec{\pi}_{1:K} \given \vec{\beta}, \alpha, \kappa, s_{1:T}$.}\\
\noindent\emph{Step 4: Sample $\phi_{1:K} \given y_{1:T}, s_{1:T}, H$.}\\
\noindent\emph{Step 5: Sample $(\alpha, \gamma, \kappa) \given s_{1:T}, \vec{\beta}, \vec{\pi}_{1:K}$.}
\vskip 0.1cm

Due to the strongly correlated nature of time-series data, resampling the latent hidden states in Step $1$, is often the most difficult since the other variables can be sampled via the Gibbs sampler once a sample of $s_{1:T}$ has been obtained. In the following section, we describe a novel efficient sampler for the latent states $s_{1:T}$ of the iHMM, and refer the reader to our appendix and \citet{TehJorBea2006, fox2008hdp} for a detailed discussion on steps for sampling variables $\alpha, \gamma, \kappa, \vec{\beta}, \vec{\pi}_{1:K}, \phi_{1:K}$.

\subsection{Infinite State Particle Gibbs Sampler}

Within the Particle MCMC framework of ~\citet{andrieu2010particle}, Sequential Monte Carlo (or particle filtering) is used as a complex, high-dimensional proposal for the Metropolis-Hastings algorithm. The Particle Gibbs sampler is a conditional SMC algorithm resulting from clamping one particle to an apriori fixed trajectory. In particular, it is a transition kernel that has $p(s_{1:T} | y_{1:T})$ as its stationary distribution. 

The key to constructing a generic, truncation-free sampler for the iHMM to resample the latent states, $s_{1:T}$, is to note that the finite number of particles in the sampler are ``localized" in the latent space to a finite subset of the infinite set of possible states. Moreover, they can only transition to finitely many new states as they are propagated through the forward pass. Thus the ``infinite" measure  $\vec{\beta}$, and ``infinite" transition matrix $\vec{\pi}$ only need to be instantiated to support the number of ``active" states (defined as being $\{1, ..., K\}$) in the state space. In the particle Gibbs algorithm, if a particle transitions to a state outside the ``active" set, the objects $\vec{\beta}$ and $\vec{\pi}$ can be lazily expanded via the stick-breaking constructions derived for both objects in \citet{TehJorBea2006} and stated in equations (2), (4) and (5). Thus due to the properties of both the stick-breaking construction and the PGAS kernel, this resampling procedure will leave the target distribution $p(s_{1:T} | y_{1:T})$ invariant. Below we first describe our infinite-state particle Gibbs algorithm for the iHMM then detail our notation (we provide further background on SMC in our supplement):
\vskip 0.1cm
\noindent\emph{Step 1: For iteration $t=1$ initialize as:}
\vspace{-0.2cm}
\begin{description2}
  \item [{(a)}] sample $s_1^i \dist q_{1}(\cdot)$, for $i \in 1,...,N$
  \item [{(b)}] initialize weights $w_1^i = p(s_1)f_{1}(y_1 | s_1)\big/q_{1} (s_1)$ for $i \in 1,...,N$
\end{description2}
\vspace{-0.1cm}
\emph{Step 2: For iteration $t>1$ use trajectory $s'_{1:T}$ from $t-1$, $\beta$, $\pi$, $\phi$, and $K$}: 
\vspace{-0.2cm}
 \begin{description2}
   \item [{(a)}] sample the index $a_{t-1}^i \dist \mathcal{C}at(\cdot |W_{t-1}^{1:N})$ of the ancestor of particle $i$ for $i \in 1,..., N-1$.
   \item [{(b)}] sample $s_t^i \dist q_{t}(\cdot \given s_{t-1}^{a_{t-1}^i})$  for $i \in 1,..., N-1$. If $s_t^i = K+1$ then create a new state using the stick-breaking construction for the HDP: 
   \begin{description2}
     \item [{(i)}] Sample a new transition probability vector $\vec{\pi}_{K+1} \sim \mathcal{D}ir(\alpha \vec{\beta})$.
     \item [{(ii) }] Use stick-breaking construction to iteratively expand $\beta \gets [\beta, \beta_{K+1}]$ as: \[\beta_{K+1}' \distiid \text{Beta}(1,\gamma), \quad\beta_{K+1} = \beta_{K+1}'\Pi_{\ell=1}^{K} (1-\beta_{\ell}').\]
     \item [{(iii)}] Expand transition probability vectors $(\vec{\pi}_k)$, $k=1,\ldots,K+1$, to include transitions to $K+1$st state via the HDP stick-breaking construction as:
     \[\vec{\pi}_{j} \gets [\pi_{j1}, \pi_{j2},\ldots, \pi_{j, K+1}], \quad \forall j=1,\ldots,K+1.\]
      where \[\vec{\pi}'_{jK+1} \sim \text{Beta}\big(\alpha_{0}\beta_K, \alpha_{0}(1-\sum_{\ell=1}^{K+1} \beta_{l})\big), \
\vec{\pi}_{jK+1} = \vec{\pi}'_{jK+1} \Pi_{\ell=1}^{K}(1-\vec{\pi}'_{j\ell}).\]
     \item [{(iv)}] Sample a new emission parameter $\phi_{K+1} \sim H$.
    \end{description2}
    \item [{(c)}] compute the ancestor weights $\tilde{w}^i_{t-1|T} = w^i_{t-1}\pi(s'_t | s^i_{t-1})$ and resample $a_t^N$ as \[\mathbb{P}(a_t^N=i) \propto \tilde{w}^i_{t-1|T}.\]
    \item [{(d)}] recompute and normalize particle weights using:
     \begin{align}
    w_{t}(s_t^i) &= \pi(s_t^i \given s_{t-1}^{a_{t-1}^i})f(y_t \given s_t^i)\slash q_{t}(s_t^i\given s_{t-1}^{a_{t-1}^i}) \nonumber \\
    W_{t} (s_t^i) &= w_{t}(s_t^i) \slash (\sum_{i=1}^{N}w_{t}(s_t^i)) \nonumber
   \end{align}
   \end{description2}
\vspace{-0.2cm}
   \emph{Step 3:} Sample $k$ with $\mathbb{P}(k=i) \propto w^{i}_{T}$ and return $s^{*}_{1:T} = s^{k}_{1:T}$.
\vskip 0.1cm   
In the particle Gibbs sampler, at each step $t$ a weighted particle system  $\{s_t^i, w_t^i\}_{i=1}^N$ serves as an empirical point-mass approximation to the distribution $p(s_{1:T})$, with the variables $a^i_{t}$ denoting the `ancestor' particles of $s^{i}_{t}$.
Here we have used $\pi(s_t | s_{t-1})$ to denote the latent transition distribution, $f(y_t | s_t)$ the emission distribution, and $p(s_1)$ the prior over the initial state $s_1$. 
\subsection{More Efficient Importance Proposal $q_t(\cdot)$}
In the PG algorithm described above, we have a choice of the importance sampling density $q_t(\cdot)$ to use at every time step.  The simplest choice is to sample from the ``prior" -- $q_t(\cdot | s_{t-1}^{a_{t-1}^i}) =  \pi(s_t^i | s_{t-1}^{a_{t-1}^i})$ -- which can lead to satisfactory performance when then observations are not too informative and the dimension of the latent variables are not too large. If we were to use the ``prior" as our importance sampling density then the time-complexity of our sampler would be strictly $\mathcal{O}(TN)$. However using the prior as importance proposal in particle MCMC is known to be suboptimal. 
In order to improve the mixing rate of the sampler, it is desirable to sample from the partial ``posterior" -- $q_{t}(\cdot \given s_{t-1}^{a_{t-1}^i}) \propto \pi(s_t^i | s_{t-1}^{a_{t-1}^i})f(y_t | s_t^i)$ -- whenever possible .

In general, sampling from the ``posterior", $q_{t}(\cdot \given s_{t-1}^{a_{t-1}^n}) \propto \pi(s_t^n | s_{t-1}^{a_{t-1}^n})f(y_t | s_t^n)$, may be impossible, but in the iHMM we can show that it is analytically tractable. To see this, note that we have lazily represented $\pi(\cdot | s_{t-1}^n)$ as a finite vector --  $[\pi_{s_{t-1}^n, 1:K}, \pi_{s_{t-1}^n, K+1}]$. Moreover, we can easily evaluate the likelihood $f(y_t^n | s_{t}^n, \phi_{1:K})$ for all $s_{t}^n \in {1, ..., K}$. However, if $s_{t}^n = K+1$, we need to compute $f(y_t^n | s_{t}^n = K+1) = \int {f(y_t^n | s_{t}^n = K+1, \phi) H(\phi)} d\phi$. If $f$ and $H$ are conjugate, we can analytically compute the marginal likelihood of the $K+1$st state, but this can also be approximated by Monte Carlo sampling for non-conjugate likelihoods -- see \citet{neal2000a} for a more detailed discussion of this argument. Thus,  we can compute $p(y_t | s_{t-1}^n) = \sum_{k=1}^{K+1} \pi(k \given s_{t-1}^n)f(y_t \given \phi_k)$ for each particle $s^n_t$ where $n \in 1, ..., N-1$. 

We investigate the impact of ``posterior" vs. ``prior" proposals in \Cref{priorposterior}. Based on the convergence of the number of states and joint log-likelihood, we can see that sampling from the ``posterior" improves the mixing of the sampler. Indeed, we see from the "prior" sampling experiments that increasing the number of particles from $N=10$ to $N=50$ does seem to marginally improve the mixing the sampler, but have found $N=10$ particles sufficient to obtain good results. However, we found no appreciable gain when increasing the number of particles from $N=10$ to $N=50$ when sampling from the ``posterior" and omitted the curves for clarity. However, it is worth noting that the PG sampler does still perform reasonably even when sampling from the ``prior" with the added advantage that it's time complexity will simply be $\mathcal{O}(TN)$.

\subsection{Improving Mixing via Ancestor Resampling}
It has been recognized that the mixing properties of the PG kernel can be poor due to path degeneracy \citep{lindsten2014particle}. A variant of PG that is presented in \citet{lindsten2014particle} attempts to address this problem for any non-Markovian state-space model with a modification -- resample a new value for the variable $a_t^N$ in an ``ancestor sampling" step at every time step, which can significantly improve the mixing of the PG kernel with little extra computation in the case of Markovian systems.

To understand ancestor sampling, for $t\geq2$ consider the reference trajectory $s'_{t:T}$ ranging from the current time step $t$ to the final time $T$. Now, artificially assign a candidate history to this partial path, by connecting $s'_{t:T}$ to one of the other particles history up until that point $\{s_{1:t-1}^i\}_{i=1}^{N}$ which can be achieved by simply assigning a new value to the variable $a_t^N \in {1, ..., N}$.  To do this, we first compute the weights:
\begin{equation}
\tilde{w}^i_{t-1|T} \equiv w^i_{t-1} \frac{p_{T}(s^i_{1:t-1}, s'_{t:T} | y_{1:T})}{p_{t-1}(s^i_{1:t-1}| y_{1:T})}, \quad i=1,...,N
\end{equation}
Then $a_t^N$ is sampled according to $\mathbb{P}(a_t^N=i) \propto \tilde{w}^i_{t-1|T}$. Remarkably, this ancestor sampling step leaves the density $p(s_{1:T}\given y_{1:T})$ invariant as shown in \citet{lindsten2014particle} for arbitrary, non-Markovian state-space models. However since the infinite HMM is Markovian, we can show the computation of the ancestor sampling weights simplifies to 
\begin{equation}
\tilde{w}^i_{t-1|T} = w^i_{t-1}\pi(s'_t | s^i_{t-1})
\end{equation}
Note that the ancestor sampling step does not change the $O(TNK)$ time complexity of the infinite-state PG sampler.

\subsection{Resampling $\vec{\pi}$, $\vec{\phi}$, $\vec{\beta}$, $\alpha$, $\gamma$, and $\kappa$}

Our resampling scheme for $\vec{\pi}$, $\vec{\beta}$, $\vec{\phi}$, $\alpha$, $\gamma$, and $\kappa$ will follow straightforwardly from this scheme in ~\citet{fox2008hdp, TehJorBea2006}. We present a review of their methods and related work in our appendix for completeness.

\section{Empirical Study}\label{sec:empirical}

In the following experiments we explore the performance of the PG sampler on both the iHMM and the sticky iHMM. Note that throughout this section we have only taken $N=10$ and $N=50$ particles for the PG sampler which has time complexity $\mathcal{O}(TNK)$ when sampling from the ``posterior", and $\mathcal{O}(TN)$ when sampling from the ``prior", compared to the time complexity of $\mathcal{O}(TK^2)$ of the beam sampler. For completeness, we also compare to the Gibbs sampler, which has been shown perform worse than the beam sampler \citep{VanSaaTeh2008a}, due to strong correlations in the latent states.

\subsection{Convergence on Synthetic Data}

To study the mixing properties of the PG sampler on the iHMM and sticky iHMM, we consider two synthetic examples with strongly positively correlated latent states. First as in \citet{VanSaaTeh2008a}, we generate sequences of length 4000 from a 4 state HMM with self-transition probability of $0.75$, and residual probability mass distributed uniformly over the remaining states where the emission distributions are taken to be normal with fixed standard deviation $0.5$ and emission means of $-2.0, -0.5, 1.0, 4.0$ for the 4 states. The base distribution, $H$ for the iHMM is taken to be normal with mean 0 and standard deviation 2, and we initialized the sampler with $K=10$ ``active" states.
\begin{figure}[t]
\vspace{-1.7cm}
    \centering
    \begin{minipage}[t]{.52\textwidth}
        \centering
        \includegraphics[width=1\linewidth]{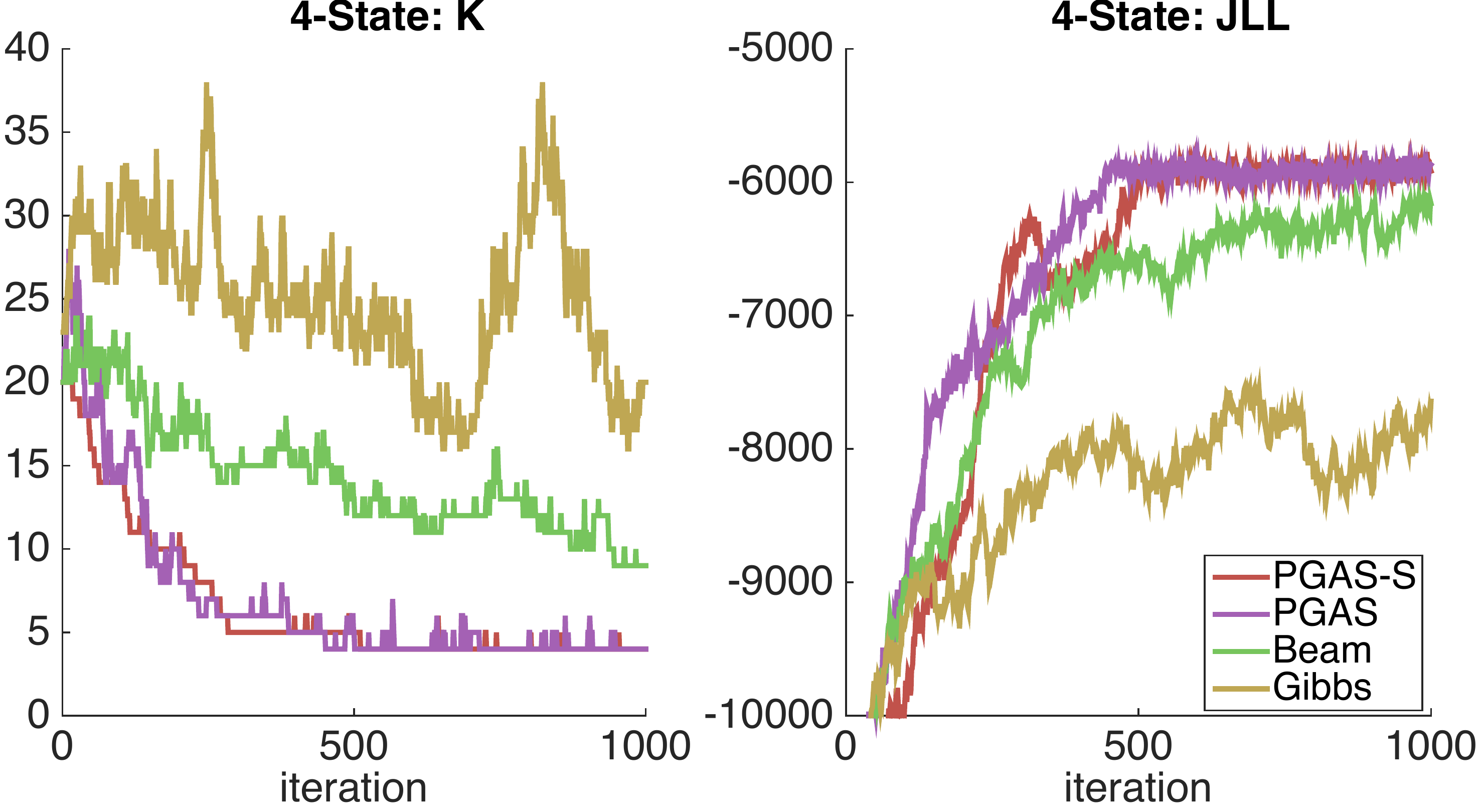}
        \caption{Comparing the performance of the PG sampler, PG sampler on sticky iHMM (PG-S), beam sampler, and Gibbs sampler on inferring data from a 4 state strongly correlated HMM. Left: Number of ``Active" States K vs. Iterations Right: Joint-Log Likelihood vs. Iterations (Best viewed in color)}
        \label{4K}
    \end{minipage}%
    ~
    \begin{minipage}[t]{0.46\textwidth}
        \centering
        \includegraphics[width=1\linewidth]{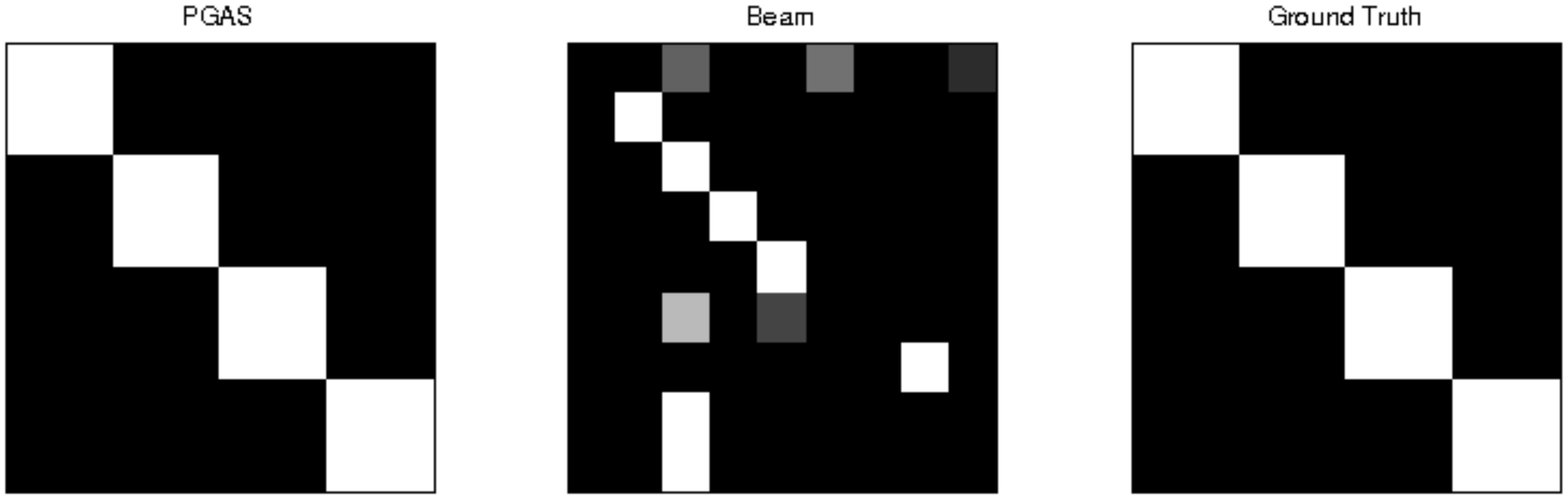}
        \vspace{-1.2cm}
        \caption{Learned Latent Transition Matrices for the PG sampler and Beam Sampler vs Ground Truth (Transition Matrix for Gibbs Sampler omitted for clarity). PG correctly recovers strongly correlated self-transition matrix, while the Beam Sampler supports extra ``spurious" states in the latent space.}
        \label{matrices}
    \end{minipage}
\end{figure}
In the 4-state case, we see in \Cref{4K} that the PG sampler applied to both the iHMM and the sticky iHMM converges to the ``true" value of $K=4$ much quicker than both the beam sampler and Gibbs sampler -- uncovering the model dimensionality, and structure of the transition matrix by more rapidly eliminating spurious ``active" states from the space as evidenced in the learned transition matrix plots in \Cref{matrices}. Moreover, as evidenced by the joint log-likelihood in \Cref{4K}, we see that the PG sampler applied to both the iHMM and the sticky iHMM converges quickly to a good mode, while the beam sampler has not fully converged within a $1000$ iterations, and the Gibbs sampler is performing poorly.

To further explore the mixing of the PG sampler vs. the beam sampler\footnote{We no longer explore the performance of the Gibbs sampler since based on our previous experiment, and extensive experimentation in \citet{VanSaaTeh2008a}, we believe the Gibbs sampler to be strictly worse than the beam sampler.} we consider a similar inference problem on synthetic data over a larger state space. We generate data from sequences of length $4000$ from a $10$ state HMM with self-transition probability of $0.75$, and residual probability mass distributed uniformly over the remaining states, and take the emission distributions to be normal with fixed standard deviation $0.5$ and means equally spaced 2.0 apart between $-10$ and $10$. The base distribution, $H$, for the iHMM is also taken to be normal with mean 0 and standard deviation 2. The samplers were initialized with $K=3$ and $K=30$ states to explore the convergence and robustness of the infinite-state PG sampler vs. the beam sampler.
\begin{figure}[!htb]
    \centering
    \begin{minipage}[t]{.48\textwidth}
        \centering
        \includegraphics[width=1\linewidth]{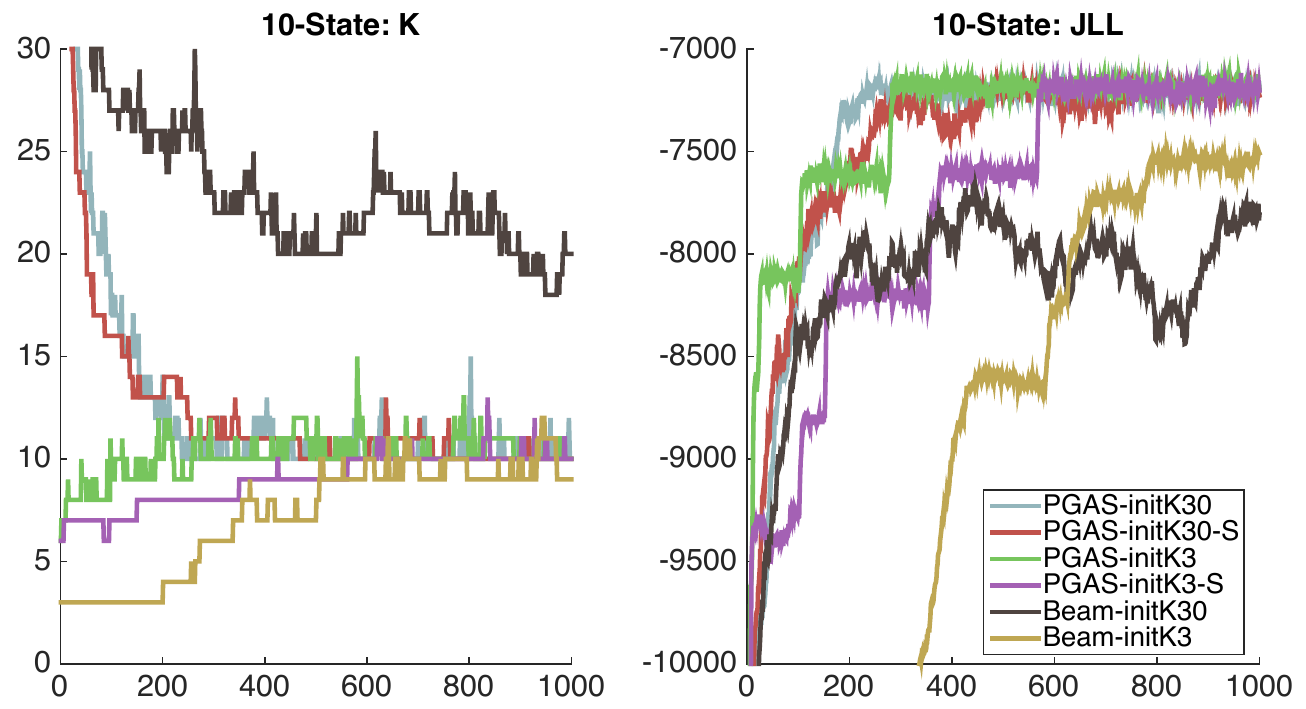}
        \caption{Comparing the performance of the PG sampler vs. beam sampler on inferring data from a 10 state strongly correlated HMM with different initializations. Left: Number of ``Active" States K from different Initial K vs. Iterations Right: Joint-Log Likelihood from different Initial K vs. Iterations}\label{fig:10statesa}
    \end{minipage}%
    ~
    \begin{minipage}[t]{0.48\textwidth}
        \centering
        \includegraphics[width=1\linewidth]{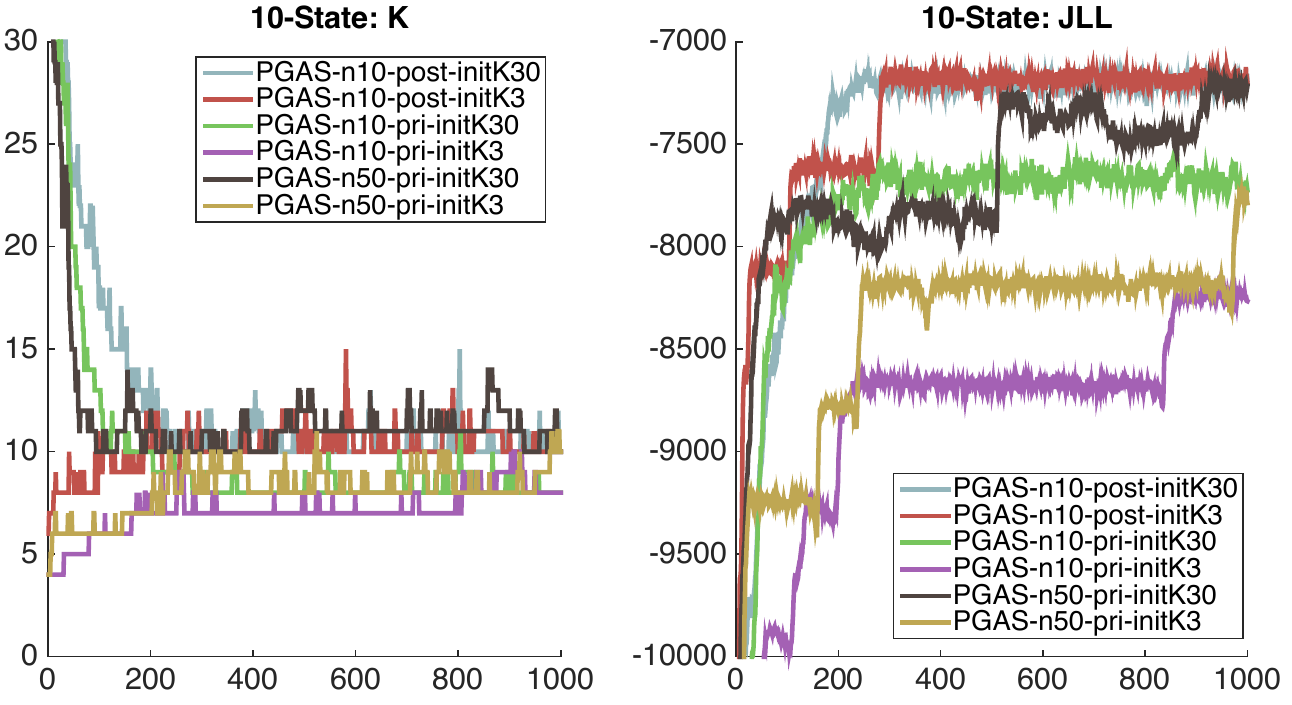}
        \caption{Influence of ``Posterior" vs. ``Prior" proposal and Number of Particles in PG sampler on iHMM. Left: Number of ``Active" States K from different Initial K, Numbers of Particles, and ``Prior"/``Posterior" proposal  vs. Iterations Right: Joint-Log Likelihood from different Initial K, Numbers of Particles, and ``Prior"/"Posterior" proposal vs. Iterations}
        \label{priorposterior}
    \end{minipage}
    \vspace{-.5cm}
\end{figure}

As observed in \Cref{fig:10statesa}, we see that the PG sampler applied to the iHMM and sticky iHMM, converges far more quickly from both ``small" and ``large" initialization of $K=3$ and $K=30$ ``active" states to the true value of $K=10$ hidden states, as well as converging in JLL more quickly. Indeed, as noted in ~\citet{fox2008hdp}, the introduction of the extra slice variables in the beam sampler can inhibit the mixing of the sampler, since for the beam sampler to consider transitions with low prior probability one must also have sampled an unlikely corresponding slice variable so as not to have truncated that state out of the space. This can become particularly problematic if one needs to consider several of these transitions in succession. We believe this provides evidence that the infinite-state Particle Gibbs sampler presented here, which does not introduce extra slice variables, is mixing better than beam sampling in the iHMM.

\subsection{Ion Channel Recordings}

For our first real dataset, we investigate the behavior of the PG sampler and beam sampler on an ion channel recording. In particular, we consider a 1MHz recording from \citet{rosenstein2013single} of a single alamethicin channel previously investigated in \citet{palla2014reversible}. We subsample the time series by a factor of 100, truncate it to be of length 2000, and further log transform and normalize it. 

\begin{figure}[!htb]
  \vspace{-2.0cm}
   \centering
   \includegraphics[trim=0cm 6cm 0cm 6cm, clip, width=.8\textwidth]{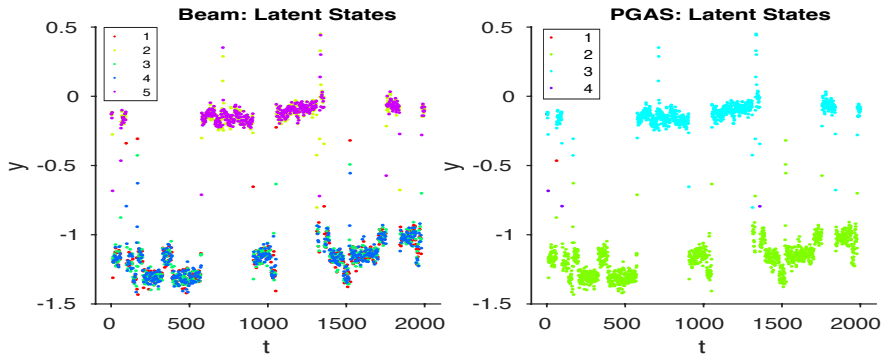}
   \vspace{-2.0cm}
   \caption{Left: Observations colored by an inferred latent trajectory using beam sampling inference. Right: Observations colored by an inferred latent state trajectory using PG inference.}
   \label{ion}
   \vspace{-0.2cm}
\end{figure}

We ran both the beam and PG sampler on the iHMM for 1000 iterations (until we observed a convergence in the joint log-likelihood). Due to the large fluctuations in the observed time series, the beam sampler infers the number of ``active" hidden states to be $K=5$ while the PG sampler infers the number of ``active" hidden states to be $K=4$. However in \Cref{ion}, we see that beam sampler infers a solution for the latent states which rapidly oscillates between a subset of likely states during temporal regions which intuitively seem to be better explained by a single state. However, the PG sampler has converged to a mode which seems to better represent the latent transition dynamics, and only seems to infer ``extra" states in the regions of large fluctuation. Indeed, this suggests that the beam sampler is mixing worse with respect to the PG sampler.

\subsection{Alice in Wonderland Data}
For our next example we consider the task of predicting sequences of letters taken from \textit{Alice's Adventures in Wonderland}. We trained an iHMM on the 1000 characters from the first chapter of the book, and tested on 4000 subsequent characters from the same chapter using a multinomial emission model for the iHMM. 
\begin{figure}[!htb]
  \vspace{-2.5cm}
   \centering
   \includegraphics[width=0.85\linewidth]{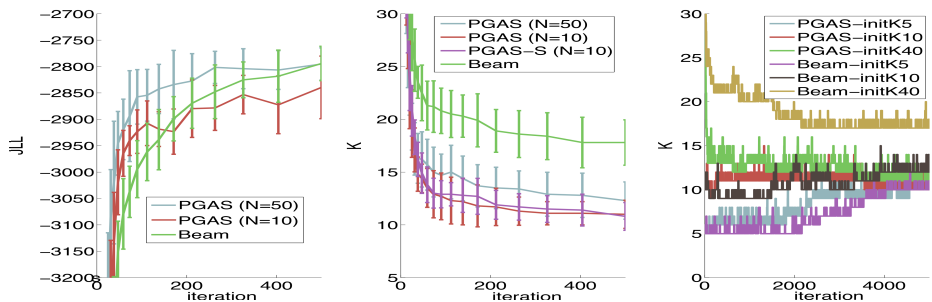}
   \vspace{-2.5cm}
   \caption{Left: Comparing the Joint Log-Likelihood vs. Iterations for the PG sampler and Beam sampler. Middle: Comparing the convergence of the ``active" number of states for the iHMM and sticky iHMM for the PG sampler and Beam sampler. Right: Trace plots of the number of states for different initializations for K.}
   \label{alice}
   \vspace{-.5cm}
\end{figure}

Once again, we see that the PG sampler applied to the iHMM/sticky iHMM converges quickly in joint log-likelihood to a mode where it stably learns a value of $K \approx 10$ as evidenced in \Cref{alice}. Though the performance of the PG and beam samplers appear to be roughly comparable here, we would like to highlight two observations. Firstly, the inferred value of $K$ obtained by the PG sampler quickly converges independent of the initialization $K$ in the rightmost of \Cref{alice}. However, the beam sampler's prediction for the number of active states $K$ still appears to be decreasing and more rapidly fluctuating than both the iHMM and sticky iHMM as evidenced by the error bars in the middle plot in addition to being quite sensitive to the initialization $K$ as shown in the rightmost plot. Based on the previous synthetic experiment (Section 4.1), and this result we suspect that although both the beam sampler and PG sampler are quickly converging to good solutions as evidenced by the training joint log-likelihood, the beam sampler is learning a transition matrix with unnecessary\big/spurious ``active" states. Next we calculate the predictive log-likelihood of the Alice in Wonderland test data averaged over 2500 different realizations and find that the infinite-state PG sampler with $N=10$ particles achieves a predictive log-likelihood of $\vec{-5918.4 \pm 123.8}$ while the beam sampler achieves a predictive log-likelihood of $\vec{-6099.0 \pm 106.0}$, showing the PG sampler applied to the iHMM and Sticky iHMM learns hyperparameter and latent variable values that obtain better predictive performance on the held-out dataset. 
We note that in this experiment as well, we have only found it necessary to take $N=10$ particles in the PG sampler achieve good mixing and empirical performance, although increasing the number of particles to $N=50$ does improve the convergence of the sampler in this instance. Given that the PG sampler has a time complexity of $\mathcal{O}(TNK)$ for a single pass, while the beam sampler (and truncated methods) have a time complexity of $\mathcal{O}(TK^2)$ for a single pass, we believe that the PG sampler is a competitive alternative to the beam sampler for the iHMM.

\section{Discussions and Conclusions}\label{sec:discussion}

In this work we derive a new inference algorithm for the iHMM using the particle MCMC framework based on the stick-breaking construction for the HDP. We also develop an efficient proposal inside PG optimized for iHMMs, to efficiently resample the latent state trajectories of the iHMM, and the sticky iHMM. The proposed algorithm is empirically compared to existing state-of-the-art inference algorithms for iHMMs, and the algorithm presented here is particularly promising because it converges more quickly and robustly to the true number of states in addition to obtaining better predictive performance on several synthetic and realworld datasets. 
Moreover, we argued that the PG sampler proposed here is a competitive alternative to the beam sampler since the time complexity of the particle samplers presented is $\mathcal{O}(TNK)$\footnote{Assuming we use the ``posterior" as the importance sampling density.} versus the $\mathcal{O}(TK^2)$ of the beam sampler.

Another advantage of the proposed method is the simplicity of the PG algorithm, which doesn't require truncation or the introduction of auxiliary variables, also making the algorithm easily adaptable to challenging inference tasks. In particular, the PG sampler can be directly applied to the sticky HDP-HMM with DP emission model considered in \cite{fox2008hdp} for which no truncation-free sampler exists. We leave this development and application as an avenue for future work.

\bibliographystyle{apalike}
\bibliography{hdphmm.bib}

\newpage
\appendix

\section{Hierarchical Dirichlet Process}
A Dirichlet process (DP), parametrized as $\text{DP}(\gamma, H)$, is a stochastic process whose realizations are countably infinite measures:
\begin{align}
G(\phi) = \sum_{k=1}^{\infty} \beta_{k} \delta_{\phi_{k}}, \quad \phi_{k} \sim H
\end{align}
over some parameter space $\Phi$. Here $H$ is the base measure defined on the space $\Phi$, while $\gamma$ is a scalar concentration parameter controlling the variability of the process around $H$ (lower $\gamma$ implies more variability). The weights, $\beta_{k}$ of the DP can be sampled via a stick-breaking construction \citep{sethuraman1991constructive}:
\begin{align}
\beta_{k}' \distiid \text{Beta}(1,\gamma), \quad\beta_{k} = \beta_{k}'\Pi_{\ell=1}^{k-1} (1-\beta_{\ell}')
\label{stick1}
\end{align}
referred to as $\vec{\beta} \sim \text{GEM}(\gamma)$. Importantly for our purposes, note the $\beta_{k}$ are defined in a purely recursive fashion.

The Hierarchical Dirichlet Process (HDP) of ~\citep{TehJorBea2006} takes a hierarchical Bayesian approach by defining multiple DP's that share one random measure that is itself drawn from a DP. This hierarchical coupling allows one to non-parametrically model individual subgroups that are generated uniquely but share some overall information. Specifically, we have that 
\begin{align}
\begin{split}
G_{0} \sim \text{DP}(\gamma, H), \ G_{k} \sim \text{DP}(\alpha, G_{0}) \quad \forall k
\end{split}
\end{align}
Here $\alpha$ controls the variability of each $G_{k}$ around the shared base measure $G_{0}$, while $H$ is the global base measure over the parameter space. By appealing to the stick-breaking construction for the DP we can express the random measures succinctly as: 
\begin{align}
\begin{split}
\beta \sim \text{GEM}(\gamma), \ 
G_{0} \sim \sum_{k'=1}^{\infty} \beta_{k'} \delta_{\phi_{k'}},  \
G_{k} = \sum_{k'=1}^{\infty} \vec{\pi}_{kk'} \delta_{\phi_{k'}}
\end{split}
\label{stick1}
\end{align}
with
\begin{align}
\begin{split}
\vec{\pi}'_{jk} \sim \text{Beta}\big(\alpha_{0}\beta_k, \alpha_{0}(1-\sum_{\ell=1}^{k} \beta_{l})\big), \
\vec{\pi}_{jk} = \vec{\pi}'_{jk} \Pi_{\ell=1}^{k-1}(1-\vec{\pi}'_{j\ell})
\end{split}
\label{stick2}
\end{align}
and $\beta_{k}$ defined as before.

\section{Particle MCMC}

The key idea of the Particle Markov Chain Monte Carlo framework (PMCMC) of ~\citet{andrieu2010particle} is that Sequential Monte Carlo (or particle filtering) is used as a complex, high-dimensional proposal for Metropolis-Hastings. The Particle Gibbs sampler results from using the conditional SMC algorithm, which clamps one particle to an apriori fixed trajectory. Crucially, the Particle Gibbs algorithm will leave the target distribution invariant (we refer the reader to the original paper for further technical details). 

First we review the construction of the SMC sampler for finite-state space models. Let $p(s_{1:T} | y_{1:T})$ denote the target density of the latent states parametrized by some $\theta \in \Theta$, with prior $p(s_1)$ over the initial state. Then let $\{s_t^i, w_t^i\}_{i=1}^N$ be a weighted particle system at time $t$ serving as an empirical point-mass approximation to the distribution $p(s_{1:T})$, with the variables $a^i_{t}$ denoting the ancestor particles of $s^{i}_{t}$. For the state-space model dynamics, we will use $\pi(s_t | s_{t-1})$ to denote the latent transition density, $f(y_t | s_t)$ the conditional likelihood, and $p(s_{1:T}, y_{1:T})$ the joint likelihood.

The algorithm is initialized by sampling $s_1^i \sim q_{1,\theta} (\cdot)$ from a proposal density and initializing the importance weights as $w_1^i = p(s_1)f_{\theta,1}(y_1 | s_1)/q_{\theta,1} (s_1)$.  We can then describe the SMC kernel on $N$ particles indexed as $i \in {1,...,N}$:

\noindent\emph{Step 1: For iteration $t=1$:}
  \begin{description2}
    \item [{(a)}] sample $s_1^i \dist q_{1,\theta}(\cdot)$
    \item [{(b)}] initialize weights $w_1^i = p(s_1)f_{\theta,1}(y_1 | s_1)/q_{\theta,1} (s_1)$
  \end{description2}
  \emph{Step 2: For iteration $t>1$:}
  \begin{description2}
    \item [{(a)}] sample the index $a_{t-1}^i \dist \mathcal{M}ult(\cdot |W_{t-1, \theta}^{1:N})$ of the ancestor of particle $i$ for $i \in 1,..., N$
    \item [{(b)}] sample $s_t^i \dist q_{t,\theta}(\cdot \given s_{t-1}^{a_{t-1}^i})$ 
    \item [{(c)}] recompute and normalize weights
    \begin{align}
    w_{t, \theta}(s_t^i) &= \pi_{\theta}(s_t^i \given s_{t-1}^{a_{t-1}^i})f_{\theta}(y_t \given s_t^i)\slash q_{t,\theta}(s_t^i\given s_{t-1}^{a_{t-1}^i}) \nonumber \\
    W_{t, \theta} (s_t^i) &= w_{t, \theta}(s_t^i) \slash (\sum_{i=1}^{N}w_{t, \theta}(s_t^i)) \nonumber
   \end{align}
  \end{description2}

The Particle Gibbs sampler is similar to the SMC sampler, but conditions on the event that one particle in the system is constrained to a reference trajectory $s'_{1:T} = (s'_1, ..., s'_T)$. This is accomplished by only resampling for $i=1,...,N-1$ in parts b) and c) above. After one pass of the conditional SMC algorithm, an entire trajectory is sampled as $\mathbb{P}(s^*_{1:T} = s^i_{1:T}) \propto w^i_T$ where $s^i_{1:T}$ is constructed by tracing the ancestors of $s^i_T$ back through the sampled trajectories. 

\section{Sampling Other Variables and Related Work}

The goal of any sampling scheme for the iHMM is to sample the variables $s_{1:T}, \vec{\beta}, \vec{\pi}_{1:K}, \phi_{1:K}, \alpha, \gamma, \kappa$. 

Building on the direct assignment sampling scheme for the HDP derived in \citet{TehJorBea2006}, the original Gibbs sampler took the approach of first marginalizing out the infinite, latent variables $\vec{\pi}$ and $\vec{\phi}$ in (6). Thus we need only explicitly resample the hidden trajectory $\bold{s}$, the base DP parameters $\beta$, and hyper parameters $\alpha$ and $\gamma$. Sampling $\beta$, $\alpha$, and $\gamma$ follows directly from the theory of the HDP, and the stick-breaking construction. To sample $s_t$ conditional on $s_{-t}, \beta, y, \alpha, H$ for $t \in {1, ..., T}$, we need to compute the conditional $p(s_t | s_{-t}, \beta, y, \alpha, H) \propto p(y_t | s_t, s_{-t}, y_{-t}, H) p(s_t | s_{-t}, \beta, \alpha)$. The first factor is simply the conditional likelihood: $p(y_t | s_t, s_{-t}, y_{-t}, H) = \int{ p(y_t | s_t, \phi_{s_{t}}) p(\phi_{s_{t}} |s_{-t}, y_{-t}, H) d\phi_{s_{t}}}$, which is easily computed when the base distribution $H$ is conjugate to the likelihood $f$. The second factor can be easily computed using the Markov property of the hidden state sequence. Since for each $t \in {1, ..., T}$ we compute $O(K)$ probabilities, the Gibbs sampler has $O(TK)$ complexity. The Gibbs sampler's is straightforwardly implemented but often suffers from slow mixing behavior since sequential data tends to be strongly correlated.

In contrast, the traditional approach for efficient inference of the hidden state trajectory in the classical, finite-state space HMM uses the forward-backwards algorithm (i.e. belief propagation) to recursively infer the hidden state trajectory in $\mathcal{O}(TK^2)$ time where $T$ is the length of the HMM and $K$ the size of the latent space. It is tempting to hope a similar type of algorithm exists for the iHMM, but it is impossible to directly apply such a message-passing approach due to the countably infinite state-space (i.e. $K$ is unbounded). However, ~\citet{VanSaaTeh2008a} circumvented this difficulty in the iHMM by introducing of a set of auxiliary slice variable $u_{1:T}$ into the model; when conditioned on $u_{1:T}$ the model becomes finite. In contrast to the original Gibbs sampling routine, the beam sampler iteratively resamples auxiliary slice variables $u$,
the trajectory $\vec{s}$, transition matrix $\pi$, shared DP measure $\beta$, and hyper-parameters $\alpha$, $\gamma$ conditioned on all other variables. This allowed ~\citet{VanSaaTeh2008a} to use dynamic programming to jointly resample the latent states. In practice, they found that their sampler mixed much faster than the naive Gibbs sampler and had average-case complexity closer to $\mathcal{O}(TK)$ for sparse transition matrices, but worst-case complexity $\mathcal{O}(TK^2)$ \citep{VanSaaTeh2008a}. 

Despite the power of the beam-sampling scheme, ~\citet{fox2008hdp} found that application of the beam sampler to the sticky iHMM, resulted in slow mixing. As noted in ~\citet{fox2008hdp}, for the beam sampler to consider transitions with low prior probability one must also have sampled an unlikely corresponding slice variable so as not to have truncated that state out of the space. Such a situation becomes increasingly problematic if one must make several of these low-probability moves, independently of whether there is strong data-dependent likelihood favoring the transition. This problem was avoided in ~\citet{fox2008hdp} by considering a fixed-order truncation of the HDP-HMM and designing a blocked Gibbs sampler to resample the finite set of latent states at the cost of introducing bias into the inference. Although the truncation affords the possibility of exploring the full set of paths unhindered by the slice variables, one must balance the trade-off between the bias and computational cost of the blocked sampler on the truncated model -- $\mathcal{O}(TK^2)$ where $K$ must be taken to be large to obtain small bias. This more complex variant of the iHMM coupled with a Dirichlet Process (DP) emission distribution achieved state-of-the-art performance on a particularly challenging speaking diarization task. Notably, the ``stickiness" helped eliminate the undesirable fast-transition behaviour characteristic of the HDP-HMM\footnote{This only requires the introduction of a single extra hyper-parameter}, and the DP emission model helped capture the complex, multimodal nature of the data. Indeed, it is worth noting that although the beam sampler can be applied to the sticky iHMM, it cannot be applied to the sticky iHMM with a nonparametric DP emission model. In fact, no exact sampler has been previously constructed for this model. 

Our resampling scheme for $\pi$, $\phi$, $\beta$, $\alpha$, $\gamma$, and $\kappa$ will follow straightforwardly from this scheme in ~\citep{VanSaaTeh2008a}, ~\citep{fox2008hdp} and ~\citep{TehJorBea2006}. We refer the reader to these works for the details on the resampling of $\alpha$, $\kappa$ and $\gamma$  since we use exactly the constructions presented there, but present a brief overview of how to sample $\pi$, $\phi$, and $\beta$.

For simplicity we will review the case of the normal iHMM where $\kappa=0$ since the introduction of $\kappa$ involves more bookkeeping but does not modify the core scheme. 
Let $n_{ij}$ denote the number of times state $i$ transitions to state $j$ in the trajectory $s_{1:T}$, and $K$ be the number of distinct states in $s_{1:T}$. Merging the infinitely many states not represented in $s$ into one state, the conditional distribution of $( \pi(1 | s_t), ... ,\pi(K | s_t), \sum_{s'=K+1}^{\infty} \pi(s'_{t+1} | s_t) )$ given its Markov blanket $\vec{s}$, $\vec{\beta}$, and $\alpha$ is $$\mathcal{D}ir(n_{s_{k}1} + \alpha\beta_{1}, ..., n_{s_{k}K} + \alpha\beta_{K}, \alpha \sum_{i=K+1}^{\infty} \beta_i)$$
To sample $\vec{\beta}$ we first introduce a set of auxiliary variables $m_{jk}$ which can be interpreted as the number of times parameter $\phi_{k}$ has been sampled in $\pi_{j}$ (sometimes these parameters are referred to as dishes in the Chinese Restaurant Franchise analogy). These parameters have conditional distributions equal to $p(m_{jk} = m | \vec{s}, \vec{\beta}, \alpha, \kappa) \propto S(n_{ij}, m) (\alpha \beta_j)^m$ where $S(\cdot, \cdot)$ denote the Stirling numbers of the first kind. As \citet{TehJorBea2006} and \citet{antoniak1974mixtures} show this gives the conditional distribution over $\vec{\beta}$ as $\mathcal{D}ir(m_{\cdot k},..., m_{\cdot K}, \gamma)$ where $m_{\cdot k} = \sum_{k'=1}^K m_{k'k'}$. 
Conditional on all other variables the $\vec{\phi}_{k}$ are independent of each other and can be easily sampled efficiently when the base distribution $H$ is conjugate to the data distribution $F$, though the assumption of conjugacy is not necessary.

\bibliographystyle{apalike}
\bibliography{hdphmm.bib}

\end{document}